\documentclass[10pt,twocolumn,letterpaper]{article}

\usepackage{iccv}
\usepackage{times}
\usepackage{epsfig}
\usepackage{graphicx}
\usepackage{amsmath}
\usepackage{amssymb}

\usepackage{color,xcolor}
\usepackage{multirow}
 
\usepackage{makecell}

\usepackage[ruled]{algorithm2e}
\usepackage{algorithmic}

\usepackage{bm}
\usepackage{dsfont}
\usepackage{booktabs}

\usepackage{color}
\usepackage{xcolor}

\usepackage{subfigure}
\usepackage{enumitem}

\usepackage[pagebackref=true,breaklinks=true,letterpaper=true,colorlinks,bookmarks=false]{hyperref}

\iccvfinalcopy 

\def\iccvPaperID{4368} 

\ificcvfinal\fi

\begin{document}

\title{I-ViT: Integer-only Quantization for Efficient Vision Transformer Inference}

\author{Zhikai Li$^{1,2}$, Qingyi Gu$^{1,}$\thanks{Corresponding author.} \\
$^1$Institute of Automation, Chinese Academy of Sciences\\
$^2$School of Artificial Intelligence, University of Chinese Academy of Sciences\\
{\tt\small \{lizhikai2020, qingyi.gu\}@ia.ac.cn}}

\maketitle
\ificcvfinal\thispagestyle{empty}\fi

\begin{abstract}
  Vision Transformers (ViTs) have achieved state-of-the-art performance on various computer vision applications. However, these models have considerable storage and computational overheads, making their deployment and efficient inference on edge devices challenging. Quantization is a promising approach to reducing model complexity, and the dyadic arithmetic pipeline can allow the quantized models to perform efficient integer-only inference. Unfortunately, dyadic arithmetic is based on the homogeneity condition in convolutional neural networks, which is not applicable to the non-linear components in ViTs, making integer-only inference of ViTs an open issue. In this paper, we propose I-ViT, an integer-only quantization scheme for ViTs, to enable ViTs to perform the entire computational graph of inference with integer arithmetic and bit-shifting, and without any floating-point arithmetic. In I-ViT, linear operations (e.g., MatMul and Dense) follow the integer-only pipeline with dyadic arithmetic, and non-linear operations (e.g., Softmax, GELU, and LayerNorm) are approximated by the proposed light-weight integer-only arithmetic methods. More specifically, I-ViT applies the proposed Shiftmax and ShiftGELU, which are designed to use integer bit-shifting to approximate the corresponding floating-point operations. We evaluate I-ViT on various benchmark models and the results show that integer-only INT8 quantization achieves comparable (or even slightly higher) accuracy to the full-precision (FP) baseline. Furthermore, we utilize TVM for practical hardware deployment on the GPU's integer arithmetic units, achieving 3.72$\sim$4.11$\times$ inference speedup compared to the FP model. Code of both Pytorch and TVM is released at \url{https://github.com/zkkli/I-ViT}.
\end{abstract}

\section{Introduction}
Vision Transformers (ViTs) have recently achieved great success on a variety of computer vision tasks \cite{han2022survey,dosovitskiy2020image,carion2020end}. Nevertheless, as compared to convolutional neural networks (CNNs), ViTs suffer from higher memory footprints, computational overheads, and power consumption, hindering their deployment and real-time inference on resource-constrained edge devices \cite{li2022patch,hou2022multi,jia2021efficient,tang2022patch}. Thus, compression approaches for ViTs are being widely researched.

\begin{figure}[t]
  \centering
  \subfigure[FasterTransformer \cite{fastertransformer}]{\includegraphics[width=0.42\linewidth]{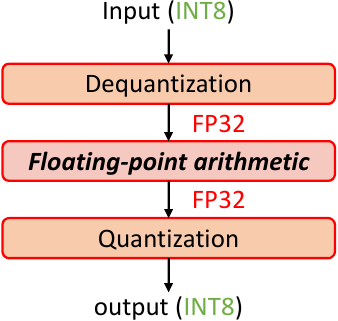}}
      \label{fig:1-a}
  \quad \quad
  \subfigure[I-ViT (ours)]{\includegraphics[width=0.42\linewidth]{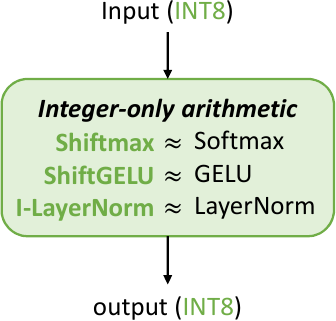}}
      \label{fig:1-b}
  \caption{Computation flows of Softmax, GELU, and LayerNorm in FasterTransformer \cite{fastertransformer} and our proposed I-ViT. I-ViT realizes the entire computational graph with integer-only arithmetic, which is more promising and practical for low-cost model deployment and efficient inference.}
  \label{Fig1}
\end{figure}

Model quantization, which reduces the representation precision of weight/activation parameters, is an effective and hardware-friendly way to improve model efficiency \cite{gholami2021survey,krishnamoorthi2018quantizing,choi2018pact,qin2020binary}. 
With the quantized low-precision parameters, previous work \cite{jacob2018quantization} presents the dyadic arithmetic pipeline to realize integer-only inference, where the quantization scaling factors are collapsed into the integer multiplication and bit-shifting in the requantization process. This can enable the quantized models to fully benefit from the fast and efficient low-precision integer arithmetic units and thus provides promising speedup effects \cite{wu2020integer,yao2021hawq}. For instance, the edge processor core in ARM Cortex-M family only support the deployment of the integer-only kernels; the recent Turing Tensor Cores in GPU server class also add support for integer logical units, and their high throughput capability enables notably lower latency compared to floating-point arithmetic.

However, the above integer-only pipeline is designed for CNNs and works under the homogeneity condition, making it only applicable to linear ($e$.$g$., Dense) or piecewise linear ($e$.$g$., ReLU) operations \cite{jacob2018quantization,yao2021hawq}. Therefore, the non-linear operations ($e$.$g$., Softmax, GELU, and LayerNorm) in ViTs cannot naively follow it. To cope with this problem, a brute-force scheme is to simply leave the non-linear operations as dequantized floating-point arithmetic, such as FasterTransformer \cite{fastertransformer} shown in Figure \ref{fig:1-b}(a). Unfortunately, this scheme makes them tolerate the inefficiency of floating-point arithmetic units, and the cut of the computational graph also introduces communication costs between integer and floating-point units, which severely limits the speedup of inference. In addition, low-cost integer-only hardware cannot meet mixed-precision computing requirements, hence one has to design heterogeneous chips by adding floating-point arithmetic units, which definitely increases the budget for model deployment.

Consequently, integer-only arithmetic for non-linear operations is significant for low-cost deployment and efficient inference. To this end, several works have attempted on language Transformer models. Fully-8bit \cite{lin2020towards} employs L1 LayerNorm to replace the non-linear arithmetic of standard deviation, and I-BERT \cite{kim2021bert} proposes integer polynomial approximations for the non-linear operations. However, such approaches are inefficient and fail to fully exploit the benefits of hardware logic. Moreover, they are developed for language models, making it infeasible to properly transfer to ViTs due to differences in data distribution.
For ViTs, FQ-ViT \cite{lin2021fq} preliminarily explores the feasibility of integer arithmetic for part of the operations ($e$.$g$., Softmax), but it is simply built on I-BERT \cite{kim2021bert} and ignores the notable GELU operation, leaving a huge gap between it and integer-only inference.
As a result, \textbf{\emph{how to accurately perform the non-linear operations of ViTs with efficient integer-only arithmetic}} remains an open issue.

In this paper, we propose I-ViT, which quantizes the entire computational graph to fill the research gap of integer-only quantization for ViTs. 
Specifically, linear operations follow the dyadic arithmetic pipeline; and non-linear operations are approximated without accuracy drop by novel light-weight integer-only arithmetic methods, where Shiftmax and ShiftGELU perform most arithmetic with bit-shifting that can be efficiently executed with simple shifters in hardware logic \cite{wang2022chip}, and I-LayerNorm calculates the square root with integer iterations instead. 

\begin{figure}[t]
    \centering
    \includegraphics[width=0.8\linewidth]{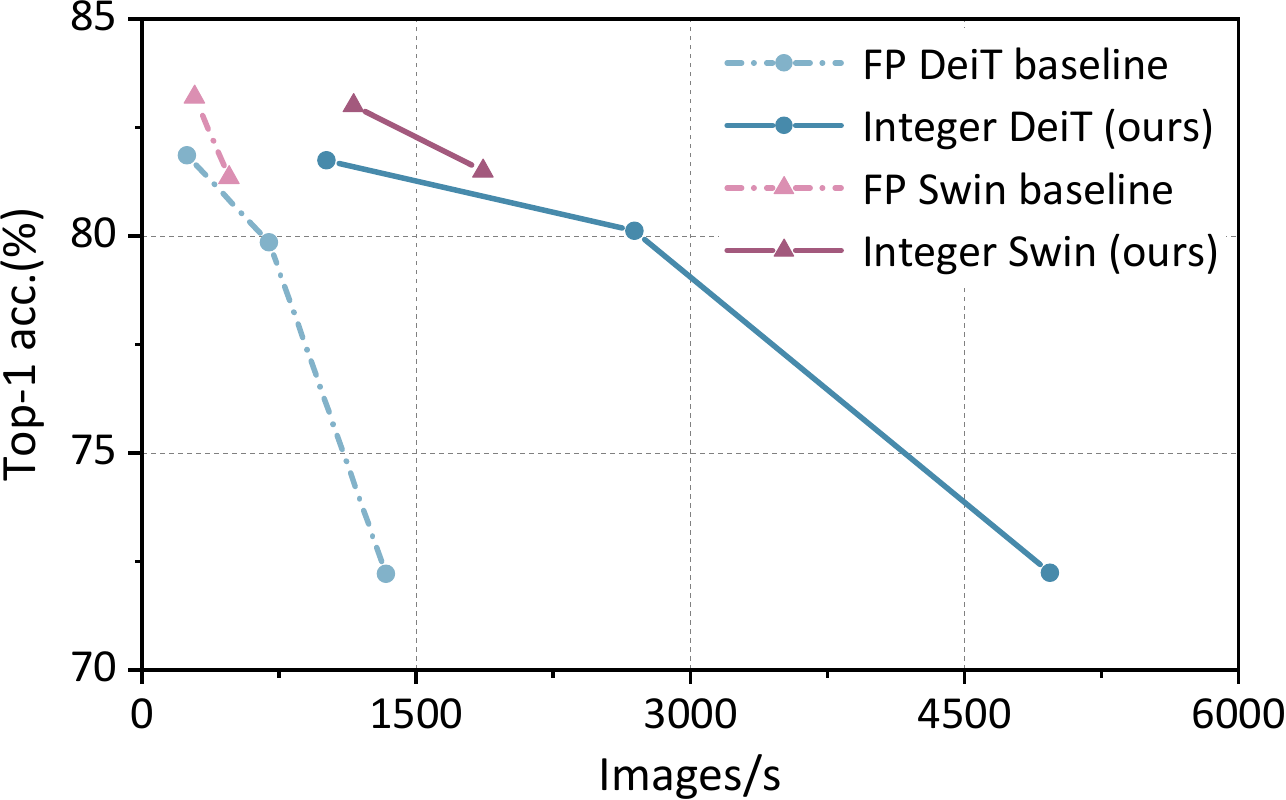}
    \caption{Accuracy-speed curves of I-ViT and the FP baseline on DeiT \cite{touvron2021training} and Swin \cite{liu2021swin}. Accuracy is evaluated on ImageNet dataset, and speed is obtained from the latency on an RTX 2080Ti GPU (batch=8). As we can see, I-ViT provides significant accelerations (3.72$\sim$4.11$\times$) while achieving similar (or even slightly higher) accuracy.}
    \label{Fig:exp1}
  \end{figure}

The main contributions are summarized as follows:

\begin{itemize}
  \item We propose I-ViT, which fully quantizes the computational graph of ViTs and allows performing the entire inference with integer arithmetic and bit-shifting, without any floating-point operations. To the best of our knowledge, this is the first work on integer-only quantization for ViTs.
  \item We propose novel light-weight integer approximations for non-linear operations (as shown in Figure \ref{fig:1-b}(b)), in particular, Shiftmax and ShiftGELU use integer bit-shifting to accomplish most arithmetic, which fully benefit from the efficient hardware logic.
  \item I-ViT is evaluated on various models for the large-scale classification task, achieving compression with similar (or even slightly higher) accuracy. Moreover, we deploy I-ViT on an RTX 2080Ti GPU using TVM\footnote{\url{https://github.com/apache/tvm}} \cite{chen2018tvm}, which accelerates the integer-only inference of ViTs with Turing Tensor Cores, achieving a 3.72$\sim$4.11$\times$ speedup over the FP model (as shown in Figure \ref{Fig:exp1}).
\end{itemize}

\section{Related Works}

\subsection{Vision Transformers}
Thanks to the global receptive fields captured by the attention mechanism, ViTs have shown superior performance on various computer vision tasks \cite{han2022survey,wu2021rethinking,han2021transformer}. 
ViT \cite{dosovitskiy2020image} is the first effort to apply transformer-based models to vision applications and achieves high accuracy than CNNs on the classification task. 
DeiT \cite{touvron2021training} introduces an efficient teacher-student strategy via adding a distillation token, reducing the time and data cost in the training phase.
Swin \cite{liu2021swin} presents shifted window attentions at various scales, which boosts the performance of ViTs. 
Furthermore, ViTs have also been applied to more complexed vision applications, such as object detection \cite{carion2020end,zhu2020deformable}, semantic segmentation \cite{chen2021pre}, and video recognition \cite{arnab2021vivit}.

Despite the promising performance, ViTs' complicated architectures with large memory footprints and computational overheads is intolerable in real-world applications \cite{hou2022multi,yang2021nvit,zhang2022minivit,li2022repq}, especially in time/resource-constrained scenarios. Thus, the compression approaches for ViTs are necessary for practical deployments.

\subsection{Model Quantization}
Model quantization, which converts the floating-point parameters to low-precision values, is a prevalent solution to compressing models in a hardware-friendly manner \cite{krishnamoorthi2018quantizing,gholami2021survey,li2022dual,zhou2017incremental}. 
Most previous works are designed to quantize CNNs. 
DoReFa \cite{zhou2016dorefa} and LQ-Net \cite{zhang2018lq} approximate the gradient propagation in quantization-aware training by straight-through estimator (STE) \cite{bengio2013estimating}. 
PACT \cite{choi2018pact} and LSQ \cite{esser2019learned,bhalgat2020lsq+} treat the activation clipping value/step size as trainable parameters and achieve promising results on low-bit quantization.
In addition, several notable works adopt more advanced quantization strategies, including non-uniform quantization \cite{li2019additive}, channel-wise quantization \cite{li2019fully}, and mixed-precision quantization \cite{wang2019haq,dong2019hawq} etc.

Recently, several quantization methods oriented to ViTs' unique structures are proposed.
Ranking loss \cite{liu2021post} is presented to maintain the correct relative order of the quantized attention map.
Q-ViT \cite{li2022q-vit} proposes differentiable quantization for ViTs, taking the quantization bit-widths and scales as learnable parameters.
PTQ4ViT \cite{yuan2021ptq4vit} proposes twin uniform quantization and uses a Hessian guided metric to evaluate different scaling factors.
FQ-ViT \cite{lin2021fq} introduces powers-of-two scale quantization and log-int quantization for LayerNorm and Softmax, respectively.
RepQ-ViT \cite{li2022repq} decouples the quantization and inference processes to address the extreme distributions of LayerNorm and Softmax activations.
PSAQ-ViT \cite{li2022patch,li2022psaq} pushes the quantization of ViTs to data-free scenarios based on patch similarity.

However, in the above approaches, all or part of the operations are performed with dequantized floating-point parameters during inference, which fails to fully use efficient low-precision arithmetic units and thus provides unsatisfactory model acceleration.

\subsection{Integer-only quantization} 

Integer-only quantization, which eliminates dequantization and enables the entire inference to be performed with integer-only arithmetic, can potentially address the above challenges. Dyadic arithmetic is proposed to perform the integer-only pipeline for CNNs \cite{jacob2018quantization,yao2021hawq}, however, it is designed for linear and piecewise linear operations based on the homogeneity condition, and thus is not applicable to non-linear operations in ViTs. 


Therefore, several studies are interested in how to achieve integer arithmetic for non-linear operations in language Transformer models. Fully-8bit \cite{lin2020towards} introduces L1 LayerNorm, which avoids the non-linearity of solving for the square root when calculating the standard deviation.
I-BERT \cite{kim2021bert} focuses on integer polynomial approximations for the non-linear operations, including Softmax, GELU, and LayerNorm.
It is worth noting that although these three component approximations can potentially enable integer-only inference of ViTs (partly verified by FQ-ViT \cite{lin2021fq}), the computation of high-order polynomials is inefficient in inference, and they are developed for language models that do not fit the data distribution of ViTs, leading to mismatched approximations.
In addition, various approximation methods that hold floating-point arithmetic are presented \cite{stevens2021softermax,zhu2020efficient}; while they lower certain computational costs, they cannot meet the demands of integer arithmetic. As a result, integer-only quantization for ViTs remains a research gap.


\begin{figure}[t]
  \centering
  \includegraphics[width=1\linewidth]{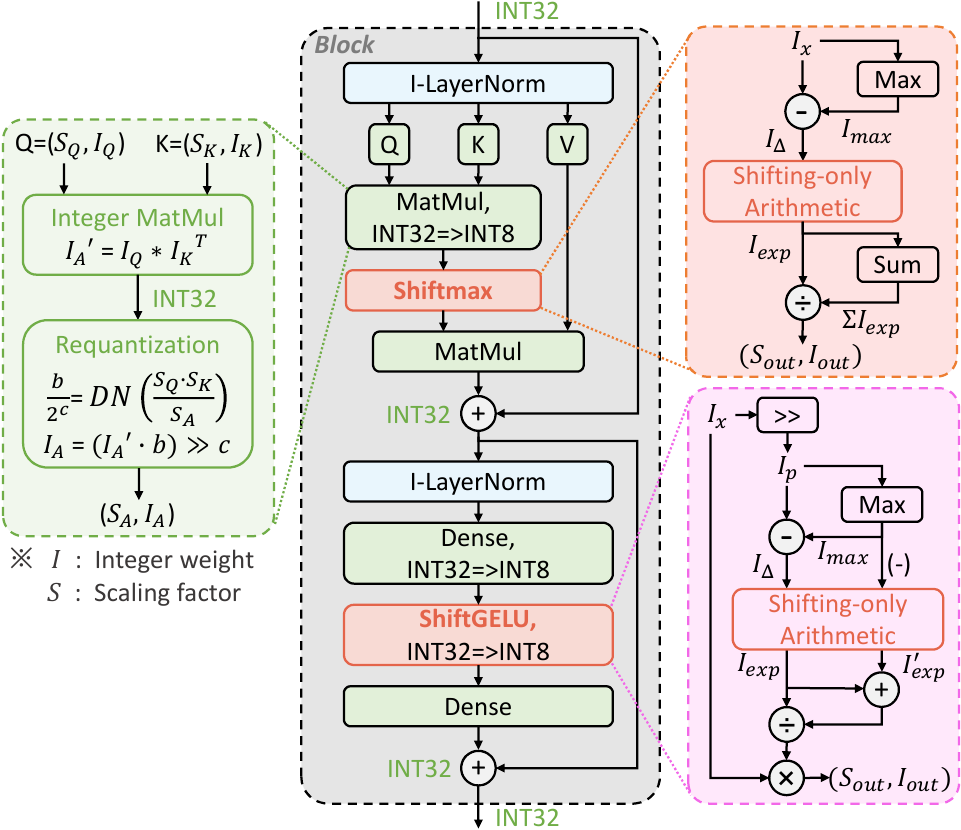}
  \caption{Overview of the proposed I-ViT. The entire computational graph is performed with integer-only arithmetic, where linear MatMul and Dense operations follow the dyadic arithmetic pipeline and the proposed Shiftmax, ShiftGELU, and I-LayerNorm accomplish the non-linear operations. Except for the labeled INT32, the remaining data streams are all INT8 precision.}
  \label{Fig:overview}
\end{figure}

\section{Methodology}

\subsection{Overview}
The overview of the proposed integer-only quantization scheme for ViTs is illustrated as Figure \ref{Fig:overview}.
The main body of ViTs is a stack of blocks, and each block is divided into a multi-head self-attention (MSA) module and a multi-layer perceptron (MLP) module, which can be formulated as follows:
\begin{align}
  \hat X & = \text{MSA}(\text{LayerNorm}(X)) + X \\
  Y & = \text{MLP}(\text{LayerNorm}(\hat X)) + \hat X
\end{align}

The MSA module learns inter-patch representations by calculating the global attention as follows: 
\begin{align}
  \text{MSA}(X) = \text{Concat}&(\text{Attn}_{1}, \text{Attn}_{2},\ldots, \text{Attn}_{h}) W^O \\
  \text{where} \;\; \text{Attn}_{i} &= \text{Softmax}\left(\frac{Q_i\cdot K_i^T}{\sqrt{d}}\right)V_i
\end{align}
where $h$ is the number of the attention heads, $d$ is the size of hidden features, and $i=1,2,\cdots,h$.
Here, $Q_i$, $K_i$, and $V_i$ are query, key, and value, respectively, and they are obtained by linear projections, $i$.$e$., $Q_i=XW_i^Q$, $K_i=XW_i^K$, $V_i=XW_i^V$.
Then the MLP module employs two dense layers and a GELU activation function to learn high-dimensional representations as follows:
\begin{equation}
  \text{MLP}(\hat X) = \text{GELU}(\hat X W_1+b_1)W_2 + b_2. \label{eq:ffn}
\end{equation}

In this work, we are interested in quantizing the entire computational graph of ViTs.
To facilitate TVM implementation, we apply the simplest \emph{symmetric uniform quantization} as follows:
\begin{equation}
	I = \left\lfloor\frac{\text{clip}(R,-m,m)}{S}\right\rceil, \; \text{where} \,\, S = \frac{2m}{2^k-1}
  \label{eq:quantization}
\end{equation}
where $R$ and $I$ denote the floating-point values and the quantized integer values, respectively, $S$ is the scaling factor of quantization, $m$ is the clipping value determined by the naive min-max method, $k$ is the quantization bit-precision, and $\lfloor\cdot\rceil$ is the round operator.

With the quantized integer values, to avoid dequantization and achieve integer-only inference, we apply the dyadic arithmetic pipeline for linear operations, as detailed in Section \ref{Sec:Dyadic}. Since the above pipeline is based on the homogeneity condition ($e$.$g$., MatMul($S_Q$·$I_Q, S_K$·$I_K$)$\equiv S_Q$·$S_K$·MatMul($I_Q, I_K$)), it is not applicable to the case of non-linearity ($e$.$g$., Softmax($S_A$·$I_A$)$\neq$$S_A$·Softmax($I_A$)). Thus, non-linear operations require accurate and efficient approximations by integer-only arithmetic. To this end, Shiftmax and ShiftGELU are proposed in this paper, which utilize efficient shifters in hardware logic to accomplish most arithmetic, and I-LayerNorm calculates the square root of the variance in an integer iterative manner. The above schemes are described in detail in Sections \ref{Sec:Shiftmax}-\ref{Sec:I-LayerNorm}, respectively.

\subsection{Dyadic Arithmetic for Linear Operations}
\label{Sec:Dyadic}

The dyadic arithmetic pipeline, which uses integer bit-shifting to efficiently realize floating-point operations of scaling factors, allows linear operations to be performed with integer-only arithmetic.
Although it is designed for CNNs \cite{jacob2018quantization,yao2021hawq}, it can also be followed for linear operations in ViTs, including Conv in the embedding layer, and MatMul and Dense in the transformer layer.

Taking MatMul as an instance, when the inputs are $Q=(S_Q,I_Q)$ and $K=(S_K,I_K)$, the output is calculated as follows:
\begin{equation}
	A' = {S_A}'\cdot {I_A}' = S_Q\cdot S_K\cdot \left(I_Q*{I_K}^T\right)
\end{equation}
where ${I_A}' = I_Q*{I_K}^T$ performs integer-only arithmetic. Following the principle of practical hardware implementation ($e$.$g$., DP4A), when the inputs $I_Q$ and $I_K$ are INT8 types, the output ${I_A}'$ is INT32 type. Thus, we need to requantize ${I_A}'$ to INT8 type as the input for the next layer, which is calculated as follows:
\begin{equation}
	I_A = \left\lfloor\frac{{S_A}'\cdot {I_A}'}{S_A}\right\rceil = \left\lfloor\frac{S_Q\cdot S_K}{S_A}\cdot \left(I_Q*{I_K}^T\right)\right\rceil
  \label{eq:I_A}
\end{equation}
where $S_A$ is the pre-calculated scaling factor of the output activation. Although the scaling factors remain floating-point values, their multiplication and division operations in Eq. \ref{eq:I_A} can be avoided by converting the rescaling to a dyadic number (DN) as follows:
\begin{equation}
	DN\left(\frac{S_Q\cdot S_K}{S_A}\right) = \frac{b}{2^c}
\end{equation}
where $b$ and $c$ are both positive integer values. In this case, the rescaling can be efficiently accomplished by integer multiplication and bit-shifting. To summarize, the integer-only arithmetic pipeline of MatMul can be denoted as follows:
\begin{equation}
	I_A = \left(b\cdot \left(I_Q*{I_K}^T\right)\right) \gg c
\end{equation}
where $\gg$ indicates right bit-shifting.

\subsection{Integer-only Softmax: Shiftmax}
\label{Sec:Shiftmax}

Softmax in ViTs translates the attention scores into probabilities, which acts on the hidden features and is calculated as follows:
\begin{equation}
    \text{Softmax}(x_i) = \frac{e^{x_i}}{\sum_j^d e^{x_j}} = \frac{e^{S_{x_i}\cdot I_{x_i}}}{\sum_j^d e^{S_{x_j}\cdot I_{x_j}}}
  \label{eq:softmax}
\end{equation}
where $i=1,2,\cdots,d$.
Due to the non-linearity, Softmax cannot follow the dyadic arithmetic pipeline discussed above, and the exponential arithmetic in Eq. \ref{eq:softmax} is typically unsupported by integer-only logic units \cite{stevens2021softermax}. To address the above issues, we propose the approximation method Shiftmax, which can utilize simple hardware logic to achieve accurate and efficient integer-only arithmetic of Softmax. First, to smooth the data distribution and prevent overflow, we restrict the range of the exponential arithmetic as follows:
\begin{equation}
	\text{Softmax}(x_i) = \frac{e^{S_{\Delta_i}\cdot I_{\Delta_i}}}{\sum_j^d e^{S_{\Delta_j}\cdot I_{\Delta_j}}} 
  = \frac{e^{S_{x_i}\cdot (I_{x_i}-I_{max})}}{\sum_j^d e^{S_{x_j}\cdot (I_{x_j}-I_{max})}}
  \label{eq:softmax_t}
\end{equation}
where $I_{max} = \max\{I_{x_1},I_{x_2},\cdots,I_{x_d}\}$. Here, $I_{\Delta_i}=I_{x_i}-I_{max}$ is a non-positive value and $S_{\Delta_i}=S_{x_i}$, and we simplify them as $I_{\Delta}$ and $S_{\Delta}$ in the following part for easier expression.

Then, we are motivated to convert the base from $e$ to $2$ to fully utilize the efficient shifters. Instead of a brute-force conversion, we perform an equivalent transformation using the base changing formula of the exponential function. Importantly, since $\log_2e$ can be approximated by binary as $(1.0111)_b$, the floating-point multiplication with it can be achieved by integer shifting as follows:
\begin{equation} 
  \begin{aligned}
  e^{S_{\Delta}\cdot I_{\Delta}} &= 2^{S_{\Delta}\cdot (I_{\Delta}\cdot \log_2e)} \\
  &\approx 2^{S_{\Delta}\cdot (I_{\Delta}+(I_{\Delta}\gg 1)-(I_{\Delta}\gg 4))} 
  \end{aligned}
  \label{eq:change_base}
\end{equation}

The power term is denoted as $S_{\Delta}$·$I_p$, which is not ensured as an integer and cannot be directly used for shifting. 
Thus, we decompose it into an integer part and a decimal part as follows:
\begin{equation}
	2^{S_{\Delta}\cdot I_p} = 2^{(-q)+S_{\Delta}\cdot (-r)} = 2^{S_{\Delta}\cdot(-r)} \gg q
  \label{eq:exp_shift}
\end{equation}
where $S_{\Delta}\text{·}(-r)\in (-1,0]$ is the decimal part, and $q$ and $r$ are both positive integer values.
For low-cost computation, we approximate $2^{S_{\Delta}\cdot(-r)}$ in range $(-1,0]$ by the linear function as follows:
\begin{equation}
  \begin{aligned}
	2^{S_{\Delta}\cdot(-r)} &\approx [S_{\Delta}\cdot(-r)]/2+1 \\
   &= S_{\Delta}\cdot[((-r)\gg 1)+I_0]
  \end{aligned}
  \label{eq:linear}
\end{equation}
where $I_0 = \lfloor 1/S_{\Delta} \rceil$.
The above completes the approximation of the numerator in Eq. \ref{eq:softmax_t}, $i$.$e$., $S_{\Delta}$·$I_{exp}\approx e^{S_{\Delta}\cdot I_{\Delta}}$, where $S_{\Delta}$ can be removed via fraction reduction since the scaling factor of the denominator obtained by summing is also $S_{\Delta}$. 
This turns Eq. \ref{eq:softmax_t} into an integer division, which is calculated with the specified output bit-precision $k_{out}$ as follows:
\begin{equation}
  \textstyle
  \begin{aligned}
  I_{out_i} &= \frac{S_{\Delta}\cdot I_{exp_i}}{S_{\Delta}\cdot \sum_j^d I_{exp_j}} \\ &=\text{IntDiv}(I_{exp_i},\sum\nolimits_j^d I_{exp_j},k_{out})
  \\ &=  \left(\left\lfloor\frac{2^M}{\sum_j^d I_{exp_j}}\right\rfloor\cdot I_{exp_i}\right) \gg (M-(k_{out}-1)) 
  \\
  S_{out_i} &= 1/2^{k_{out}-1}
  \end{aligned}
  \label{eq:IntDiv}
\end{equation}
where IntDiv$(I_1,I_2,k)$ implements the integer division function, and $I_1$, $I_2$, and $k$ are integer dividend, integer divisor and output bit width, respectively. Here, $M$ is a sufficiently large integer, and $S_{out_i}$·$I_{out_i}$\footnote{$S_{out}$ is the scaling factor for the $k_{out}$-bit symmetric quantization with $m\approx$ 1.} can approximate the result of Softmax$(x_i)$.

\begin{algorithm}[t]
	\caption{Integer-only Softmax: \textbf{Shiftmax}}
	\label{alg:shiftmax}
	\SetAlgoLined
	\KwIn{
  \quad \quad $I_{in}$ : Integer input \\
  \quad \quad \quad \quad $\;\;\;$ $S_{in}$ : Input scaling factor \\
  \quad \quad \quad \quad $\;\;$ $k_{out}$ : Output bit-precision
  }

	\KwOut{
  $\;\,\,$ $I_{out}$ : Integer output \\
  \quad \quad \quad \quad $\;\,$ $S_{out}$ : Output scaling factor \\
    }
  
  \vskip 0.075in
  \SetKwFunction{FMain}{ShiftExp}
    \SetKwProg{Fn}{Function}{:}{}
    \Fn{\FMain{$I,S$}}{
        $ I_p \leftarrow I+(I\gg 1)-(I\gg 4) $;  \hspace*{\fill}{{\color{gray}$\triangleright$ $I\cdot\log_2e$}}

        $I_0 \leftarrow \lfloor 1/S \rceil $;

        $q \leftarrow \lfloor I_p/(-I_0) \rfloor $; \hspace*{\fill}{{\color{gray}$\triangleright$ Integer part}}

        $r \leftarrow -(I_p-q\cdot(-I_0))$; \hspace*{\fill}{{\color{gray}$\triangleright$ Decimal part}}

        $I_b \leftarrow ((-r)\gg 1)+I_0$; \hspace*{\fill}{{\color{gray}$\triangleright$ Eq. \ref{eq:linear}}}
        
        $I_{exp} \leftarrow I_b \ll (N-q)$; \textcolor{red}{\footnotemark{}} \hspace*{\fill}{{\color{gray}$\triangleright$ Eq. \ref{eq:exp_shift}}}

        $S_{exp} \leftarrow S/(2^N)$;  

        \textbf{return} $(I_{exp},S_{exp})$; 
        \hspace*{\fill}{{\color{gray}$\triangleright$ $S_{exp}\cdot I_{exp}\approx e^{S\cdot I}$}}
    }
    \textbf{End Function}
    \vskip 0.075in

    \SetKwFunction{FMain}{Shiftmax}
    \SetKwProg{Fn}{Function}{:}{}
    \Fn{\FMain{$I_{in},S_{in},k_{out}$}}{
        $ I_{\Delta} \leftarrow I_{in} - \max(I_{in}) $; \hspace*{\fill}{{\color{gray}$\triangleright$ Eq. \ref{eq:softmax_t}}}

        $(I_{exp},S_{exp}) \leftarrow \texttt{ShiftExp}(I_{\Delta},S_{in})$;

        $(I_{out},S_{out}) \leftarrow \texttt{IntDiv}(I_{exp},\sum I_{exp},k_{out})$; 
        
        \hspace*{\fill}{{\color{gray}$\triangleright$ Eq. \ref{eq:IntDiv}}}

        \textbf{return} $(I_{out},S_{out})$;  
        
        \hspace*{\fill}{{\color{gray}$\triangleright$ $I_{out}\cdot S_{out}\approx\text{Softmax}(I_{in}\cdot S_{in})$}}
    }
    \textbf{End Function}
\end{algorithm}
\footnotetext{To avoid too small values after right shifting, we first have a $N$-bit left shifting.}

The integer-only flow of Shiftmax is summarized in Algorithm \ref{alg:shiftmax}.
Instead of complex second-order polynomial approximations \cite{kim2021bert}, Shiftmax performs all arithmetic with bit-shifting, except for one integer subtraction, summation, and division, which significantly improves computational efficiency.
In addition, only Eqs. \ref{eq:change_base} and \ref{eq:linear} are mathematically approximated, while all others are equivalent transformations, which ensures the accuracy of Shiftmax.

\subsection{Integer-only GELU: ShiftGELU}
\label{Sec:ShiftGELU}

GELU is the non-linear activation function in ViTs, which, from the study \cite{hendrycks2016gaussian}, can be approximated by a sigmoid function as follows:
\begin{equation}
  \begin{aligned}
	\text{GELU}(x) &= x\cdot \frac{1}{\sqrt{2\pi}}\int_{-\infty}^xe^{-t^2/2}dt \\ 
  &\approx x\cdot \sigma(1.702x) \\ &= S_x\cdot I_x\cdot \sigma(S_x\cdot 1.702I_x)
  \end{aligned}
  \label{eq:GELU}
\end{equation}

Thus, the challenge becomes the realization of the sigmoid function's integer-only arithmetic. First, 1.702 can be approximated by binary as $(1.1011)_b$, thus 1.702$I_x$ can be achieved by integer shifting, $i$.$e$., $I_p=I_x+(I_x\gg 1)+(I_x\gg 3)+(I_x\gg 4)$. Then, we equivalently transform the sigmoid function as follows:
\begin{equation}
  \begin{aligned}
  \sigma(S_x\text{·}I_p) &= \frac{1}{1+e^{-S_x\cdot I_p}} \\ &= \frac{e^{S_x\cdot I_p}}{e^{S_x\cdot I_p}+1} \\
  & = \frac{e^{S_x\cdot(I_p-I_{max})}}{e^{e^{S_x\cdot(I_p-I_{max})}}+e^{S_x\cdot(-I_{max})}}
\end{aligned}
  \label{eq:sigmoid}
\end{equation}
where, interestingly, the numerator is in exact correspondence with the numerator of Eq. \ref{eq:softmax_t}, thus the two implementations are identical. 
After that, the integer approximation of GELU is done by following the integer division in Eq. \ref{eq:IntDiv} and then multiplying it with $S_x$·$I_x$.

\begin{algorithm}[t]
	\caption{Integer-only GELU: \textbf{ShiftGELU}}
	\label{alg:shiftGELU}
	\SetAlgoLined
	\KwIn{
  \quad \quad $I_{in}$ : Integer input \\
  \quad \quad \quad \quad $\;\;\;$ $S_{in}$ : Input scaling factor \\
  \quad \quad \quad \quad $\;\;$ $k_{out}$ : Output bit-precision
  }

	\KwOut{
  $\;\,\,$ $I_{out}$ : Integer output \\
  \quad \quad \quad \quad $\;\,$ $S_{out}$ : Output scaling factor \\
    }
    \vskip 0.075in

    \SetKwFunction{FMain}{ShiftGELU}
    \SetKwProg{Fn}{Function}{:}{}
    \Fn{\FMain{$I_{in},S_{in},k_{out}$}}{
        $ I_p \leftarrow I_{in}+(I_{in}\gg 1)+(I_{in}\gg 3)+(I_{in}\gg 4) $; 

        \hspace*{\fill}{{\color{gray}$\triangleright$ $1.702I$}}

        $ I_{\Delta} \leftarrow I_p - \max(I_p) $;

        $(I_{exp},S_{exp}) \leftarrow \texttt{ShiftExp}(I_{\Delta},S_{in})$;

        $(I_{exp}',S_{exp}') \leftarrow \texttt{ShiftExp}(-\max(I_p),S_{in})$;

        $(I_{div},S_{div}) \leftarrow \texttt{IntDiv}(I_{exp},I_{exp}+I_{exp}',k_{out})$;
        
        \hspace*{\fill}{{\color{gray}$\triangleright$ Eq. \ref{eq:sigmoid}}}

        $(I_{out},S_{out}) \leftarrow (I_{in}$·$I_{div}, S_{in}$·$S_{div})$;

        \textbf{return} $(I_{out},S_{out})$;  
        
        \hspace*{\fill}{{\color{gray}$\triangleright$ $I_{out}\cdot S_{out}\approx \text{GELU}(I_{in}\cdot S_{in})$}}
    }
    \textbf{End Function}
\end{algorithm}

Algorithm \ref{alg:shiftGELU} shows the integer-only flow of ShiftGELU. Except for a few fundamental arithmetic operations, ShiftGELU utilizes shifters in hardware logic to perform all other arithmetic and thus enables the efficient inference of ViTs. Furthermore, compared to the second-order polynomial method that only approximates for a specific interval \cite{kim2021bert}, the approximation of ShiftGELU works on the entire domain of definition, which can potentially provide higher accuracy and robustness.

\subsection{Integer-only LayerNorm: I-LayerNorm}
\label{Sec:I-LayerNorm}
LayerNorm in ViTs normalizes the input in the hidden feature dimension as follows:
\begin{equation}
	\text{LayerNorm}(x) = \frac{x-\text{Mean}(x)}{\sqrt{\text{Var}(x)}} \cdot \gamma + \beta
  \label{eq:layernorm}
\end{equation}
where $\gamma$ and $\beta$ are linear affine factors.
In contrast to BatchNorm that holds fixed parameters from training and can be folded during inference, LayerNorm needs to dynamically compute statistics ($i$.$e$., mean and standard deviation) in the inference phase. The integer arithmetic units allow straightforward calculation of the mean and variance of the data, yet they fail to support the square root arithmetic for obtaining the standard deviation \cite{lin2020towards}. Thus, we improve the light-weight integer iterative approach \cite{crandall2001prime} via bit-shifting as follows:
\begin{equation}
 \begin{aligned}
	I_{i+1} &= (I_i+\lfloor\text{Var}(I_x)/I_i\rfloor)/2 \\ &= (I_i+\lfloor\text{Var}(x)/I_i\rfloor)\gg 1
\end{aligned}
  \label{eq:iterative}
\end{equation}
where $I_i$ is the result of the $i$-th iteration, and $I_0$ is initialized as $2^{\lfloor\text{bit}(\text{Var}(I_x))/2\rfloor}$.
The naive stopping criterion for the iterations is $I_{i+1}\geq I_i$, which unfortunately cannot guarantee a constant latency. We experimentally find that 10 iterations can achieve most convergence, thus we modify the stopping criterion to the iteration counts to facilitate hardware implementation.

\section{Experiments}
We evaluate I-ViT in both accuracy on the large-scale classification task and latency on the practical hardware to fully demonstrate the superiority, and I-ViT can accelerate 3.72$\sim$4.11$\times$ over the FP model while achieving similar (or even slightly higher) accuracy. Besides the \emph{FP baseline}, I-ViT is compared end-to-end with the following methods:
\begin{itemize}[itemsep=1pt,topsep=0pt,parsep=0pt]
  \item \emph{FasterTransformer} \cite{fastertransformer}: leaving non-linear operations as floating-point arithmetic.
  \item \emph{I-BERT} \cite{kim2021bert}: approximating non-linear operations with integer second-order polynomials.
\end{itemize}

Notably, several methods achieve integer inference for only part of the operations, making it impractical to perform end-to-end comparisons, hence we evaluate the following individual components in ablation studies.
\begin{itemize}[itemsep=1pt,topsep=0pt,parsep=0pt]
  \item \emph{L1 LayerNorm} in Fully-8bit \cite{lin2020towards}: using L1 norm to replace the calculation of standard deviation.
  \item \emph{LIS} in FQ-ViT \cite{lin2021fq}: using Log-Int-Softmax, which builds opon I-BERT and adds the logarithmic function.
\end{itemize}


\subsection{Accuracy Evaluation}
\label{sec:accuracy}

\begin{table*}
  \scriptsize
  \centering
  \caption{Accuracy and latency results on various model benchmarks. Here, accuracy is evaluated on ImageNet dataset, and latency is evaluated on an RTX 2080Ti GPU (batch=8). Compared to the FP baseline, I-ViT, which quantizes the entire computational graph and enables integer-only inference on Turing Tensor Cores, can achieve similar or even slightly higher accuracy and provides a significant 3.72$\sim$4.11$\times$ speedup. In addition, I-ViT consistently outperforms existing works FasterTransformer \cite{fastertransformer} and I-BERT \cite{kim2021bert} in terms of both accuracy and latency.}
  \label{tab:exp1}
  \begin{tabular}{c|cccc|cc|cc}
    \toprule
    Model & Method & Bit-prec. & Size (MB) & Int.-only & Top-1 Acc. (\%) & Diff. (\%) & Latency (ms) & Speedup \\
    \midrule
    \multirow{4.5}*{ViT-S} & Baseline & FP32 & 88 & $\times$ & 81.39 & - & 11.5 & $\times$1.00 \\
    \cmidrule{2-9}
    & FasterTransformer \cite{fastertransformer} & INT8 & 22 & $\times$ & 81.07 & {\color[HTML]{CB0000}-0.32} & 3.26 & $\times$3.53 \\
    & I-BERT \cite{kim2021bert} & INT8 & 22 & $\checkmark$ & 80.47 & {\color[HTML]{CB0000}-0.92} & 3.05 & $\times$3.77 \\
    & I-ViT (ours) & INT8 & 22 & $\checkmark$ & \textbf{81.27} & \textbf{{\color[HTML]{CB0000}-0.12}} & \textbf{2.97} & \textbf{{\color[HTML]{009901}$\times$3.87}} \\
    \midrule
    \multirow{4.5}*{ViT-B} & Baseline & FP32 & 344 & $\times$ & 84.53 & -  & 32.6  & $\times$1.00 \\
    \cmidrule{2-9}
    & FasterTransformer \cite{fastertransformer} & INT8 & 86 & $\times$ & 84.29 & {\color[HTML]{CB0000}-0.24} & 8.51 & $\times$3.83 \\
    & I-BERT \cite{kim2021bert} & INT8 & 86 & $\checkmark$ & 83.70 & {\color[HTML]{CB0000}-0.83} & 8.19 & $\times$3.98 \\
    & I-ViT (ours) & INT8 & 86 & $\checkmark$ & \textbf{84.76} & \textbf{{\color[HTML]{009901}+0.23}} & \textbf{7.93} & \textbf{{\color[HTML]{009901}$\times$4.11}}  \\
    \midrule
    \multirow{4.5}*{DeiT-T} & Baseline & FP32 & 20 & $\times$ & 72.21 & - & 5.99 & $\times$1.00 \\
    \cmidrule{2-9}
    & FasterTransformer \cite{fastertransformer} & INT8 & 5 & $\times$ & 72.06 & {\color[HTML]{CB0000}-0.15} & 1.74 & $\times$3.45 \\
    & I-BERT \cite{kim2021bert} & INT8 & 5 & $\checkmark$ & 71.33 & {\color[HTML]{CB0000}-0.88} & 1.66 & $\times$3.61 \\
    & I-ViT (ours) & INT8 & 5 & $\checkmark$ & \textbf{72.24} & \textbf{{\color[HTML]{009901}+0.03}} & \textbf{1.61} & \textbf{{\color[HTML]{009901}$\times$3.72}}  \\
    \midrule
    \multirow{4.5}*{DeiT-S} & Baseline & FP32 & 88 & $\times$ & 79.85 & - & 11.5 & $\times$1.00 \\
    \cmidrule{2-9}
    & FasterTransformer \cite{fastertransformer} & INT8 & 22 & $\times$ & 79.66 & {\color[HTML]{CB0000}-0.19} & 3.26 & $\times$3.53 \\
    & I-BERT \cite{kim2021bert} & INT8 & 22 & $\checkmark$ & 79.11 & {\color[HTML]{CB0000}-0.74} & 3.05 & $\times$3.77 \\
    & I-ViT (ours) & INT8 & 22 & $\checkmark$ & \textbf{80.12} & \textbf{{\color[HTML]{009901}+0.27}} & \textbf{2.97} & \textbf{{\color[HTML]{009901}$\times$3.87}}  \\
    \midrule
    \multirow{4.5}*{DeiT-B} & Baseline & FP32 & 344 & $\times$ & 81.85 & - & 32.6 & $\times$1.00 \\
    \cmidrule{2-9}
    & FasterTransformer \cite{fastertransformer} & INT8 & 86 & $\times$ & 81.63 & {\color[HTML]{CB0000}-0.22} & 8.51 & $\times$3.72 \\
    & I-BERT \cite{kim2021bert} & INT8 & 86 & $\checkmark$ & 80.79 & {\color[HTML]{CB0000}-1.06} & 8.19 & $\times$3.88 \\
    & I-ViT (ours) & INT8 & 86 & $\checkmark$ &  \textbf{81.74} & \textbf{{\color[HTML]{CB0000}-0.11}}  & \textbf{7.93} & \textbf{{\color[HTML]{009901}$\times$4.11}}  \\
    \midrule
    \multirow{4.5}*{Swin-T} & Baseline & FP32 & 116 & $\times$ & 81.35 & - & 16.8 & $\times$1.00 \\
    \cmidrule{2-9}
    & FasterTransformer \cite{fastertransformer} & INT8 & 29 & $\times$ & 81.06 & {\color[HTML]{CB0000}-0.29} & 4.55 & $\times$3.69 \\
    & I-BERT \cite{kim2021bert} & INT8 & 29 & $\checkmark$ & 80.15 & {\color[HTML]{CB0000}-1.20} & 4.40 & $\times$3.82 \\
    & I-ViT (ours) & INT8 & 29 & $\checkmark$ & \textbf{81.50} & \textbf{{\color[HTML]{009901}+0.15}} & \textbf{4.29} & \textbf{{\color[HTML]{009901}$\times$3.92}}  \\
    \midrule
    \multirow{4.5}*{Swin-S} & Baseline & FP32 & 200 & $\times$ & 83.20 & - & 27.8 & $\times$1.00 \\
    \cmidrule{2-9}
    & FasterTransformer \cite{fastertransformer} & INT8 & 50 & $\times$ & 83.04 & {\color[HTML]{CB0000}-0.34} & 7.35 & $\times$3.78 \\
    & I-BERT \cite{kim2021bert} & INT8 & 50 & $\checkmark$ & 81.86 & {\color[HTML]{CB0000}-1.34} & 7.13 & $\times$3.90 \\
    & I-ViT (ours) & INT8 & 50 & $\checkmark$ & \textbf{83.01} & \textbf{{\color[HTML]{CB0000}-0.19}} & \textbf{6.92} & \textbf{{\color[HTML]{009901}$\times$4.02}}  \\

  \bottomrule
\end{tabular}
\end{table*}

\textbf{Implementation Details:} I-ViT is evaluated on various popular models, including ViT \cite{dosovitskiy2020image}, DeiT \cite{touvron2021training}, and Swin \cite{liu2021swin} on ImageNet (ILSVRC-2012) \cite{krizhevsky2012imagenet} dataset for the large-scale image classification task. The pre-trained models are all obtained from timm\footnote{\url{https://github.com/rwightman/pytorch-image-models}} library.
First, we use Eq. \ref{eq:quantization} to quantize the weights of the pre-trained FP model for the initialization of I-ViT. Then, we perform quantization-aware fine-tuning using naive STE \cite{bengio2013estimating} to recover the accuracy. 
The optimizer we adopt is AdamW \cite{loshchilov2018fixing}, and the search space of the learning rate is [2e-7, 5e-7, 1e-6, 2e-6].
The above implementations are done on PyTorch\footnote{\url{https://github.com/pytorch/pytorch}}, and the model inference details ($e$.$g$., bit-shifting) follow the TVM implementation to ensure consistent accuracy with the TVM deployment.

Table \ref{tab:exp1} reports the accuracy results of I-ViT and various baselines on multiple benchmark models on ImageNet dataset.
Although I-ViT reduces the bit-precision of the parameters and enables integer-only inference, it maintains comparable accuracy, even slightly more than the FP baseline, which adequately demonstrates the effectiveness and robustness of the proposed approximation schemes.
For instance, DeiT-S obtained by I-ViT achieves 80.12\% Top-1 accuracy with 8-bit integer-only inference, which is even 0.27\% higher than the FP baseline.
In addition, I-ViT is consistently superior to FasterTransformer \cite{fastertransformer} and I-BERT \cite{kim2021bert}, and in particular, the naive application of I-BERT to ViTs suffers from mismatched approximations, making the results far from satisfactory. For Swin-S, I-BERT results in a noticeable 1.34\% accuracy drop, while I-ViT still offers high robustness.

\begin{table*}
  \scriptsize
  \centering
  \caption{Ablation studies of accuracy and latency of Shiftmax, ShiftGELU, and I-LayerNorm. Latency is evaluated on an RTX 2080Ti GPU (batch=8). Replacing ($\rightarrow$) Shiftmax and ShiftGELU with second-order polynomial approximations \cite{kim2021bert} leads to lower accuracy and higher latency, and I-LayerNorm suffers from non-trivial accuracy loss due to the mismatch in the data distribution.}
  \label{tab:ablation}
  \begin{tabular}{c|cc|cc|cc}
    \toprule
    Model & Method & Shifting-oriented & Top-1 Acc. (\%) & Diff. (\%) & Latency (ms) & Diff. (ms) \\
    \midrule
    \multirow{5.5}*{DeiT-B} & I-ViT(ours) & $\checkmark$ & \textbf{81.74} & - & 7.93 & - \\
    \cmidrule{2-7}
    & Shiftmax $\rightarrow$ Poly. \cite{kim2021bert} & $\times$ &  81.62 & {\color[HTML]{CB0000}-0.12} & 8.04 & {\color[HTML]{CB0000}+0.11} \\  
    & ShiftGELU $\rightarrow$ Poly. \cite{kim2021bert} & $\times$ & 80.88 & {\color[HTML]{CB0000}-0.86} & 8.10 & {\color[HTML]{CB0000}+0.17} \\
    & Shiftmax $\rightarrow$ LIS \cite{lin2021fq} & $\times$ & 81.66  & {\color[HTML]{CB0000}-0.08} & 8.05 & {\color[HTML]{CB0000}+0.12} \\
    & I-LayerNorm $\rightarrow$ L1 LayerNorm \cite{lin2020towards} & - & 79.25  & {\color[HTML]{CB0000}-2.49} & \textbf{7.91} & {\color[HTML]{009901}-0.02} \\
    \midrule
    \multirow{5.5}*{Swin-S} & I-ViT(ours) & $\checkmark$ & \textbf{83.01} & - & 6.92 & - \\
    \cmidrule{2-7}
    & Shiftmax $\rightarrow$ Poly. \cite{kim2021bert} & $\times$ & 82.79 & {\color[HTML]{CB0000}-0.22} & 7.02 & {\color[HTML]{CB0000}+0.10} \\
    & ShiftGELU $\rightarrow$ Poly. \cite{kim2021bert} & $\times$ & 82.10 & {\color[HTML]{CB0000}-0.91} & 7.08 & {\color[HTML]{CB0000}+0.16} \\
    & Shiftmax $\rightarrow$ LIS \cite{lin2021fq} & $\times$ & 82.89  & {\color[HTML]{CB0000}-0.12} & 7.03 & {\color[HTML]{CB0000}+0.11} \\
    & I-LayerNorm $\rightarrow$ L1 LayerNorm \cite{lin2020towards} & - & 79.69  & {\color[HTML]{CB0000}-3.32} & \textbf{6.90} & {\color[HTML]{009901}-0.02}  \\
    
  \bottomrule
\end{tabular}
\end{table*}

\subsection{Latency Evaluation}
\label{sec:latency}

\textbf{Implementation Details:} We deploy I-ViT on an RTX 2080Ti GPU using TVM to measure the real hardware latency. First, we use TVM to build and compile the same model as PyTorch, followed by the auto-tuning to optimize the computational schedule, and then we perform the end-to-end latency tests. Note that although the GPU is not an integer-only hardware, depending on the DP4A instructions, I-ViT can perform efficient integer-only inference on its Turing Tensor Cores.
Since ViT \cite{dosovitskiy2020image} and DeiT \cite{touvron2021training} have the same model structure in the inference process, ViT enjoys the same acceleration as DeiT.

The latency results of I-ViT on an RTX 2080Ti GPU (batch=8) are also shown in Table \ref{tab:exp1}. FasterTransformer \cite{fastertransformer}, which leaves non-linear operations as floating-point arithmetic and cannot be deployed on integer-only hardware, produces disappointing acceleration effects.
In the case of DeiT-T and DeiT-S quantization, it only accelerates the model by 3.45$\times$ and 3.53$\times$, respectively.
Note that the disappointing acceleration stems not only from the inefficiency of the floating-point arithmetic units, but also from the data interaction overheads between the integer and floating-point arithmetic units, since the integer results from the previous layers need to be passed to the floating-point units and returned later.
In contrast, I-BERT \cite{kim2021bert} and I-ViT can achieve integer-only inference by utilizing the integer arithmetic units of Turing Tensor Cores. Importantly, compared to I-BERT, our proposed I-ViT makes fuller use of the efficient shifters in hardware logic and thus has a more advantageous 3.72$\sim$4.11$\times$ speedup.
For instance, for DeiT-B with 32.6ms latency at FP baseline, the integer inference latencies of the quantized models obtained by I-BERT and I-ViT are 8.19ms and 7.93ms, respectively, with the latter being 0.26ms faster.
Moreover, from the results, I-ViT is more effective in accelerating more computationally-intensive models.

\begin{figure}[t]
  \centering
  \includegraphics[width=0.85\linewidth]{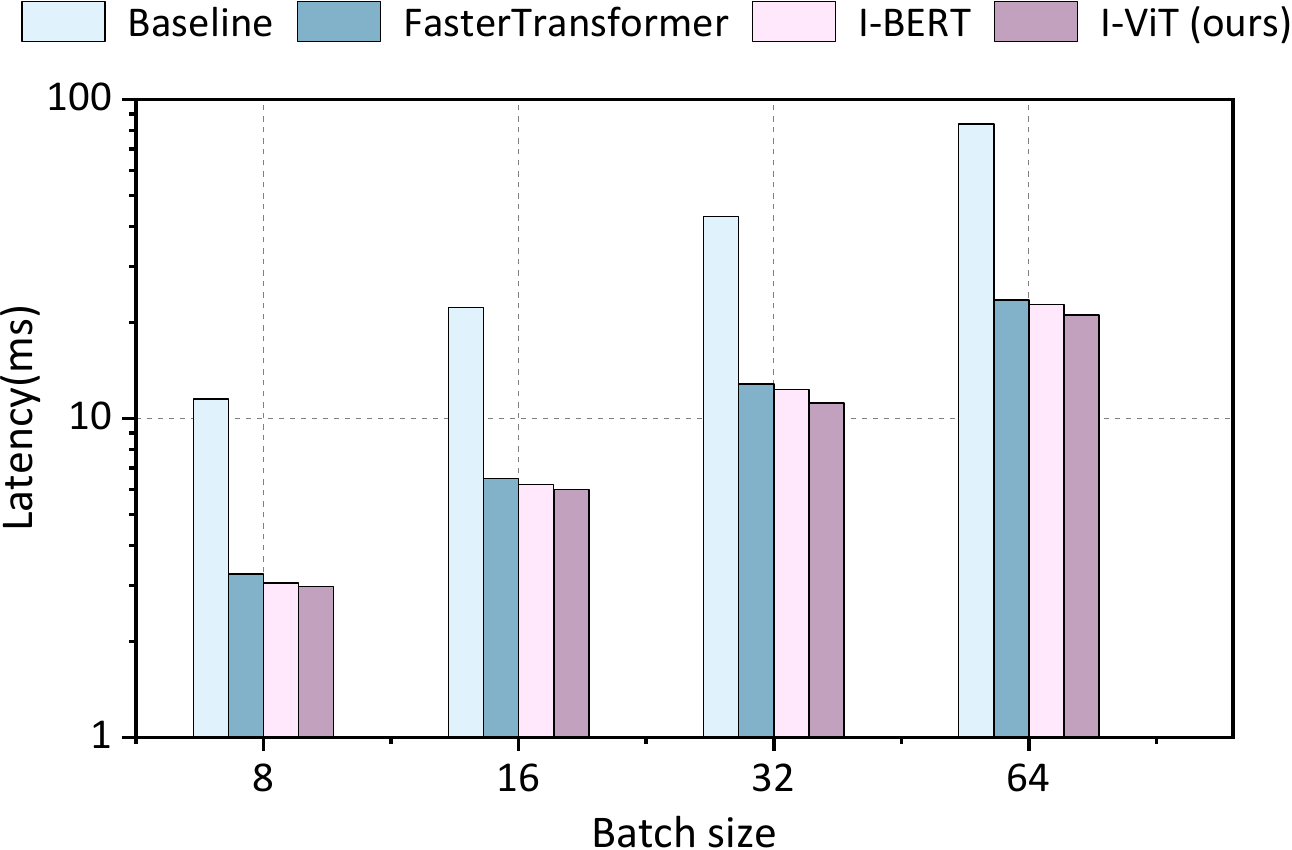}
  \caption{Latency results of DeiT-S \cite{touvron2021training} evaluated on an RTX 2080Ti GPU with various batch sizes. I-ViT maintains a constant acceleration effect for the same model architecture at various batch sizes.}
  \label{Fig:exp2}
\end{figure}

\subsection{Ablation Studies}

Here, we perform ablation studies for comparison with the second-order polynomial approximations in I-BERT \cite{kim2021bert}, LIS in FQ-ViT \cite{lin2021fq}, and L1 LayerNorm in Fully-8bit \cite{lin2020towards}, and the results are shown in Table \ref{tab:ablation}.
Due to the differences in data distribution of ViTs and language models, replacing Shiftmax and ShiftGELU with the polynomial approximations results in severe accuracy degradation, with performance losses of 0.95\% and 1.15\% in the quantization of DeiT-B and Swin-S, respectively.
In particular, polynomial GELU that only approximates for the specific interval is not applicable to ViTs and thus has most contribution in the accuracy degradation.
For instance, polynomial GELU reduces the Top-1 accuracy by 0.86\% and 0.91\% compared to ShiftGELU in the quantization of DeiT-B and Swin-S, respectively.
It is also worth mentioning that the proposed schemes are shifting-oriented arithmetic and can thus benefit more from the efficient hardware logic, while the second-order polynomial approximations lack this advantage.
LIS also encounters the above problems, since it is simply built on top of I-BERT.
For L1 LayerNorm, although it simplifies the computation to achieve faster speed, its low approximation capability leads to non-trivial accuracy loss.


In addition, we also evaluate the latency of DeiT-S with various batch sizes, as shown in Figure \ref{Fig:exp2}. It can be seen that I-ViT is robust to the batch size and can maintain a constant acceleration effect.
Also, it should be highlighted that despite the significant speedup on the RTX 2080Ti GPU that provides an evident strength of I-ViT, both the software support of TVM and the hardware support of Turing Tensor Cores are not optimal. For instance, there is no full parallelism after increasing the batch size in both FP and quantized cases, $i$.$e$., increasing the batch size results in a corresponding increase in latency. Therefore, it is believed that deploying I-ViT on dedicated hardware ($e$.$g$., FPGAs) will further enhance the acceleration potential.



\section{Conclusions}
In this paper, we propose I-ViT, which is the first integer-only quantization scheme for ViTs to the best of our knowledge.
I-ViT quantizes the entire computational graph to enable the integer-only inference, where linear operations follow the dyadic arithmetic pipeline; and non-linear operations are performed by the proposed novel light-weight integer-only approximation methods. In particular, Shiftmax and ShiftGELU perform most arithmetic with bit-shifting, which can fully benefit from the efficient hardware logic.
Compared to the FP baseline, I-ViT achieves similar (or even slightly higher) accuracy on various benchmarks. In addition, we utilize TVM to deploy I-ViT on an RTX 2080Ti GPU, whose Turing Tensor Cores can accelerates the integer-only inference of ViTs, achieving a 3.72$\sim$4.11$\times$ speedup over the FP model.

In the future, we will consider deploying I-ViT on dedicated integer-only hardware ($e$.$g$., FPGAs) to obtain better acceleration performance. Furthermore, we also plan to extend I-ViT to more complex vision tasks ($e$.$g$., object detection and semantic segmentation).

\section*{Acknowledgement}
This work was supported in part by the National Key Research and Development Program of China under Grant 2022ZD0119402; in part by the National Natural Science Foundation of China under Grant 62276255.

{\small
\bibliographystyle{ieee_fullname}
\bibliography{egbib}
}

\end{document}